# A Hierarchical Approach to Designing Approximate Reasoning-Based Controllers for Dynamic Physical Systems

Hamid Berenji*, Yung-Yaw Chen†, Chuen-Chien Lee‡, Jyh-Shing Jang§ S. Murugesan¶
Artificial Intelligence Research Branch, Mail Stop: 244-17
NASA Ames Research Center, Moffett Field, CA 94035

## Abstract

This paper presents a new technique for the design of approximate reasoning-based controllers for dynamic physical systems with interacting goals. In this approach, goals are achieved based on a hierarchy defined by a control knowledge base and remain highly interactive during the execution of the control task. The approach has been implemented in a rule-based computer program which is used in conjunction with a prototype hardware system to solve the *cart-pole balancing* problem in realtime. It provides a complementary approach to the conventional analytical control methodology, and is of substantial use where a precise mathematical model of the process being controlled is not available.

## Introduction and Motivation

Expert human controllers often perform superbly under conditions of uncertainty and imprecision using mainly approximate reasoning. They select *control actions* based on a quick assessment of the process which they are controlling. Control theorists have successfully dealt with a large class of control problems by mathematically modeling the process and solving these analytical models to generate control actions. However, the analytical models tend to become complex and infeasible to use, especially for large, intricate systems. The non-linear behavior of many practical systems makes the analytical approach even more difficult, sometimes impossible.

Starting with Mamdani and Assilian[Mamdani 75], who based their work on Zadeh's pioneering work on fuzzy set theory [Zadeh 65], an alternative method in design of controllers has been proposed. This technique, generally known as fuzzy control, has experienced much success, especially in recent years. These controllers mimic the performance of human expert operators by encoding their knowledge in terms of *linguistic control rules* which may contain fuzzy labels (e.g., HOT, MEDIUM, SMALL). Among the successful applications of this theory are the guidance control of subway trains in the city of Sendai in Japan [Yasunobu 85] and cement kiln control [Ostergaard 77]. A recent survey of this field has been provided by [Lee 90b]. In general, these controllers have been especially effective for systems with a single goal. In this paper, we provide a new method for designing approximate reasoning-based controllers which can achieve conjunctive and interacting goals. We compare our method with the state feedback control, a popular approach in modern digital control.

This comparative study is made using computer simulation and a hardware implementation of a cart-pole balancing system which represents a typical nonlinear system. This interesting problem has served as a basis for study by many connectionist works (e.g., [Widrow 87]) and control theorists (e.g., [Shaefer 66]). Learning of the control process for pole balancing has been studied by Michie and Chambers [Michie 68], Selfridge, Sutton, and Barto [Selfridge 85], [Barto 83], and by Lee [Lee 90a, Lee &Berenji 89]. In this learning research, the objective has been to write a program which can learn to keep the pole balanced.

The organization of this paper is as follows. With a brief overview of approximate reasoning-based control, we introduce a new method for designing controllers for dynamic physical systems. We then apply this method to the cart-pole balancing problem. The results of our simulations and hardware tests are discussed next. Finally, we contrast the performance of a controller based on our new approach with a conventional analytical controller.

*Sterling Federal Systems

†Dept. of Electrical Engineering, National Taiwan University, Taipei, Taiwan, R.O.C.

‡Dept. of Electrical Engineering and Computer Science, University of California, Berkeley, 94720

§Dept. of Electrical Engineering and Computer Science, University of California, Berkeley, 94720

¶ISRO Satellite Center, Banglore 560017, India

## Approximate Reasoning-Based Controllers

A difficulty in employing AI techniques in real-time control is how to handle *imprecision* in the knowledge expressed by human expert operators. Fuzzy set theory provides a facility to express the imprecise knowledge by using *linguistic variables* [Zadeh 75]. We have argued elsewhere about the importance of handling different types of uncertainty in AI systems (e.g., [Berenji 88a], [Berenji 88b]).

The basic idea in fuzzy control centers around the *labeling* process, in which the reading of a sensor is translated into a label as done by human operators. For example, in the context of controlling a nuclear reactor [Bernard 88], an observed *reactor period* (i.e., the rate of rise of the power) might be classified as *too short, short,* or *negative*. It is important to note that the transition between the labels are continuous rather than abrupt. This means that a reactor's period of 90 seconds might be termed *too short* to degree 0.2, *short* to degree 1.0, and *negative* to degree 0.0 [Bernard 88]. A similar concept is used in our experiment: an angular position of say 5 degrees might be called *Positive* to a degree of .8 and *Zero* (i.e., a label used to describe very small angles) to degree of 0.2. This idea of *partial matching* plays an important role in fuzzy control, and is related to the concept of a membership function used in fuzzy set theory where the boundary of a set is not sharp and the *degree of membership* specifies how strongly an element belongs to a set.

The knowledge base of an approximate reasoning-based controller is a collection of *linguistic control rules* which are described using *linguistic variables*[Zadeh 75]. For example,

IF $X$ is $A$ and $Y$ is $B$ THEN $Z$ is $C$

is a linguistic control rule where $X$ and $Y$ are sensor readings from the plant and $Z$ corresponds to the output (i.e., the recommended action). $A$, $B$, and $C$ are linguistic values such as LARGE, POSITIVE, etc. which are represented by membership functions (usually in triangular or trapezoidal forms). When particular values for $X$ and $Y$ are sensed, then these values are matched against the membership functions of $A$ and $B$ respectively. As a result of this matching, the degree that each precondition is satisfied will be known. Since sensor readings usually trigger several control rules at the same time, a *conflict resolution* strategy is needed. A *Max-Min compositional rule of inference*, as explained below, is commonly used.

Assume that we have the following two rules:

Rule 1: IF $X$ is $A_1$ and $Y$ is $B_1$ THEN $Z$ is $C_1$
Rule 2: IF $X$ is $A_2$ and $Y$ is $B_2$ THEN $Z$ is $C_2$

Now, if we have $x_1$ and $y_1$ as the sensor readings for fuzzy variables $X$ and $Y$, then their *truth values* are represented by $\mu_{A_1}(x_1)$ and $\mu_{B_1}(y_1)$ respectively for Rule 1, where $\mu_{A_1}$ represents the membership function for $A_1$. Similarly for Rule 2, we have $\mu_{A_2}(x_1)$ and $\mu_{B_2}(y_1)$ as the truth values of the preconditions. Then the *strength* of Rule 1 can be calculated by:

$$\alpha_1 = \mu_{A_1}(x_1) \wedge \mu_{B_1}(y_1).$$

Similarly for Rule 2:

$$\alpha_2 = \mu_{A_2}(x_1) \wedge \mu_{B_2}(y_1).$$

The effect of the strength of Rule 1 on its conclusion is calculated by:

$$\mu_{C_1'}(\omega) = \alpha_1 \wedge \mu_{C_1}(\omega),$$

and for Rule 2:

$$\mu_{C_2'}(\omega) = \alpha_2 \wedge \mu_{C_2}(\omega).$$

This means that as a result of reading sensor values $x_1$ and $y_1$, Rule 1 is recommending a control action with $\mu_{C_1'}(w)$ as its membership function and Rule 2 is recommending a control action with $\mu_{C_2'}(w)$ as its membership function. The conflict-resolution process then produces

$$\mu_C(\omega) = \mu_{C_1'}(\omega) \vee \mu_{C_2'}(\omega) = [\alpha_1 \wedge \mu_{C_1}(\omega)] \vee [\alpha_2 \wedge \mu_{C_2}(\omega)]$$

where $\mu_C(\omega)$ is a pointwise membership function for the combined conclusion of Rule 1 and Rule 2. The $\wedge$ and $\vee$ operators in above are defined to be the min and max functions respectively [Mamdani 75]. The result of this last operation ($\mu_C(\omega)$) has to be translated (*defuzzified*) to a single value. This necessary operation produces a nonfuzzy control action that best represents the membership function of an inferred fuzzy control action. The Center Of Area (COA) method (see [Lee 90b]) can be used here. Assuming a discrete universe, we have

$$Z^* = \frac{\sum_{j=1}^{n} \omega_j * \mu_C(\omega_j)}{\sum_{j=1}^{n} \mu_C(\omega_j)}$$

where $n$ is the number of quantization levels of the output.

## Hierarchical Control and Conjunctive Goal Achievement

In this section we develop a method for designing controllers which (a) obey a hierarchical process in focusing attention on a particular goal at each time instance, and (b) can achieve interacting goals simultaneously. This discussion is related in many ways to recent AI planning research where integrated planning-execution-control architectures are being explored (e.g., [Drummond 89, Bresina 90]). In this section, we present a brief discussion of our method in the general context of approximate reasoning and in Section , we demonstrate the use of this technique in the domain of cart-pole balancing. The method includes the following steps:

1. Let $G = \{g_1, g_2, ...g_n\}$ be the set of goals that system should achieve and maintain. Notice that for $n = 1$ (i.e., no interacting goals), the problem becomes simpler and may be handled using the earlier methods in fuzzy control (e.g., see [Mamdani 75]).

2. Let $G = p(G)$ where $p$ is a function which assigns priorities among the goals. We assume that such a function can be obtained in a particular domain. In many control problems, it is possible to specifically assign priorities to the goals. For example, in the simple problem of balancing a pole on the palm of a hand and also moving the pole to a pre-determined location, it is possible to do this by first keeping the pole as vertical as possible and then gradually moving to the desired location. Although these goals are highly interactive (i.e., as soon as we notice that the pole is falling, we temporarily set aside the other goal of moving to the desired location), we still can assign priorities fairly well.

3. Let $U = \{u_1, u_2, ..., u_n\}$ where $u_i$ is the set of input control parameters related to achieving $g_i$.

4. Let $A = \{a_1, a_2, ..., a_n\}$ where $a_i$ is the set of linguistic values used to describe the values of the input control parameters in $u_i$.

5. Let $C = \{c_1, c_2, ..., c_n\}$ where $c_i$ is the set of linguistic values used to describe the values of the output $Z$.

6. Acquire the rule set $R_1$ of approximate control rules directly related to the highest priority goal. These rules are in the general form of

    IF $u_1$ is $a_1$ THEN $Z$ is $c_1$.

7. For $i = 2$ to $n$, subsequently form the rule sets $R_i$. The format of the rules in these rule sets is similar to the ones in the previous step except that they include aspects of *approximately achieving the previous goal*:

    IF $g_{i-1}$ *is approximately achieved* and $u_i$ is $a_i$
    THEN $Z$ is $c_i$.

The approximate achievement of a goal in step 7 of the above algorithm refers to holding the goal parameters within smaller boundaries. The interactions among the goal $g_i$ and goal $g_{i-1}$ are handled by forming rules which include more preconditions in the left hand side. For example, let us assume that we have acquired a set of rules $R_1$ for keeping a pole vertical in the palm of a hand. In writing the second rule set $R_2$ for moving to a pre-specified location, aspects of approximately achieving $g_1$ should be combined with control parameters for achieving $g_2$. For example, a precondition such as *the pole is almost balanced* can be added while writing the rules for moving to a specific location. A fuzzy set operation known as *Concentration* [Zadeh 72] can be used here to systematically obtain a more focused membership functions for the parameters which represent the achievement of previous goals. By definition, *Concentration* is a unary operation which, when applied to a fuzzy set $A$, results in a fuzzy subset of $A$ in such a way that reduction in higher grades of membership is much less than the reduction in lower grades of membership. In other words, by concentrating a fuzzy set, members with low grades of membership will have even lower grades of memberships and hence the fuzzy set becomes more concentrated. A common concentration operator is to square the membership function:

$$\mu_{CON(A)}(y) = \mu_A^2(y)$$

and a typical concentration operator is the term *Very* which is also a *Linguistic Hedge* [Zadeh 72]. For example, the result of applying the operator *Very* on a fuzzy label *Small* is a new precondition *Very Small*.

### The Cart-Pole balancing problem

In the cart-pole balancing problem, a pole is hinged to a motor-driven cart which moves on rail tracks to its right or its left. The pole has only one degree of freedom (rotation about the hinge point). The task of a controller in this system is to keep the pole balanced within a certain small range of cart positions on the rail.

We now apply the method developed in the last section to this problem:

1. Identify the goal set:
   $G=$\{position the cart at the location $x_G$ on the track, keep the pole balanced\}.

2. Assign goal priorities:
   $g_1=$ keep the pole balanced
   $g_2=$ position the cart at the location $x_0$ on the track.

3. Identify the control parameters:
   Four state variables are used to describe the system status at each time step, and one variable represents the force applied to the cart. These are:

   $\theta$ : angle of the pole with respect to the vertical line
   $\dot{\theta}$ : angular velocity of pole
   $x$ : horizontal position of the cart on the rail
   $\dot{x}$ : velocity of the cart
   $F$ : amount of force applied to the cart to move it toward the left or the right.

   We categorize these parameters as the following:
   $u_1 = \{\theta, \dot{\theta}\}$ is the set of parameters related to keeping the pole vertically balanced
   $u_2 = \{x, \dot{x}\}$ is the set of parameters related to horizontal position control.

4. Identify the linguistic values for each input control parameter:
   Three labels are used to linguistically define the value of the four state variables: Positive, Zero, and Negative. Figure 1(a) illustrates the memberships of these linguistic terms. Hence,
$a_1$ = {Positive, Zero, Negative}.
$a_2$ = {Positive, Zero, Negative}[1].

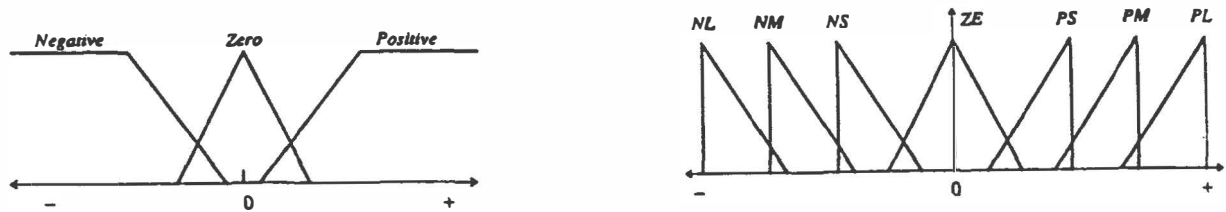

Figure 1: (a)- Three qualitative levels for $\theta, \dot{\theta}, x$, and $\dot{x}$, (b)- Seven qualitative levels for $F$

5. Identify the linguistic values for the output:
For force F, we use seven fuzzy labels: Negative-Small, Negative-Medium, Negative-Large, Zero, Positive-Small, Positive-Medium, and Positive-Large. Figure 1(b) illustrates the membership functions associated with these labels. Hence,
$c_1$ = {Positive-Large, Positive-Medium, ... , Negative-Large}
$c_2$ = {Positive-Large, Positive-Medium, ... , Negative-Large}.

6. Acquire rules for highest priority goal:
Nine control rules are used to keep the pole vertically balanced. These rules are listed in Appendix A. An example is:

IF $\theta$ is Positive and $\dot{\theta}$ is Zero THEN F is Positive-Small.

7. Form lower priority rule sets:
Approximately achieving the first goal in this problem refers to keeping the pole almost balanced. Assuming this has been achieved, we form the rule set for achieving a new cart position. For example,

IF the pole is *almost* balanced and the cart is on the right side of the desired location $x_0$,
THEN push the cart to the right[2].

The pole is assumed to be *almost balanced* if $\theta$ is *Very Small* and $\dot{\theta}$ is *Very Small*. As defined earlier, *Very Small* can be assumed to be the *Concentrated* form of the label *Small*. However, for simplicity, we use a triangular membership function with a much smaller base. This process results in more complex rules with 4 preconditions:

IF $\theta$ is *Very Small* and $\dot{\theta}$ is *Very Small* and $x$ is *Positive* and $\dot{x}$ is *Positive*,
THEN $F$ is *Positive-Medium*.

---
[1] Although the qualitative labels used are the same, different scales may be used for each control parameters.

[2] At first, this might seem to contradict our intuition to push the cart to the right when it is already on the right side of the desired position $x_0$. However, we push the cart to the right hard enough so that the pole starts to fall to the left. The subsequent attempt to keep the pole balanced will move the cart toward the center.

Four rules of the above form are used to perform cart position control. These rules are also listed in the Appendix A.

POLE is a rule-based program written in the C language and serves as the controller in this problem. It consists of only 13 rules, nine of which are used to control the angular position and the others are used to control the position of the cart. The format of the rules is simple, each one having two or four preconditions and one consequent. The main reason for the simplicity of these rules is that they are *linguistic control rules*, and the terms in the preconditions can cover a large class of sensor readings, each to a different degree.

A characteristic of the POLE program, as well as some other systems based on linguistic control, is that at any instant of time, more than one linguistic rule might be ready to fire. In this case, POLE performs conflict-resolution using the heuristic *Max-Min rule of composition* explained in Section .

## Simulations and Experiments

In this section, the performance of POLE is compared with a State Feedback Controller (SFC). SFC is one of the modern control techniques which uses a control law $u = -kx$. $u$ is the input variable of the physical system, which is a real number in single-input systems; $x$ is the state variable which is an $n$-element column vector; $k$ is the $n$-element row vector of feedback gains. The SFC formulation is based on the state space representation of the controlled system. The equations governing the cart-pole system are given in the Appendix $B^3$.

**Simulation-Based Comparison:** We first tested the performance of these controllers using computer simulation. A set of 7 poles of different lengths and weights were used. The length of these poles varied between 0.5 and 2 meters and their weights varied between 0.05 and 2.0 Kilograms. We use the notation (Pole-#, Length(m), Weight(Kg)). The poles had the following characteristics: (Pole-1, 1.0, .1), (Pole-2, .5, .05), (Pole-3, 1.0, .05), (Pole-4, .5, .025), (Pole-5, 1.0, .5), (Pole-6, 1.0, 1.0), and (Pole-7, 1.0, 2.0).

---
[3] Due to space limitations, we avoid describing the lengthy process of modeling and system identification which was required to design the SFC controller

Table 1: Comparison of the fuzzy controller (FC) and the State Feedback Controller (SFC)

|  | Pole-1 | | Pole-2 | | Pole-6 | |
| --- | --- | --- | --- | --- | --- | --- |
|  | FC | SFC | FC | SFC | FC | SFC |
| Max. $\theta$ overshoot (degrees) | .33 | 1.00 | .34 | 1.11 | .25 | .63 |
| Max. $\theta$ undershoot (degrees) | .87 | 2.29 | .73 | 2.41 | .38 | 2.52 |
| $\theta$ settling time (seconds) | 3.5 | 4.2 | 3.00 | 4.8 | 5.3 | 3.00 |
| Max. $z$ overshoot (cm) | 14.7 | 17.1 | 8.8 | 16.2 | 19.5 | 21.6 |
| Max. $z$ undershoot (Cm) | .8 | 1.7 | .5 | 2.9 | 1.6 | 0.0 |
| $z$ settling time (seconds) | 38.2 | 4.6 | 41.9 | 6.5 | 45.9 | 2.8 |

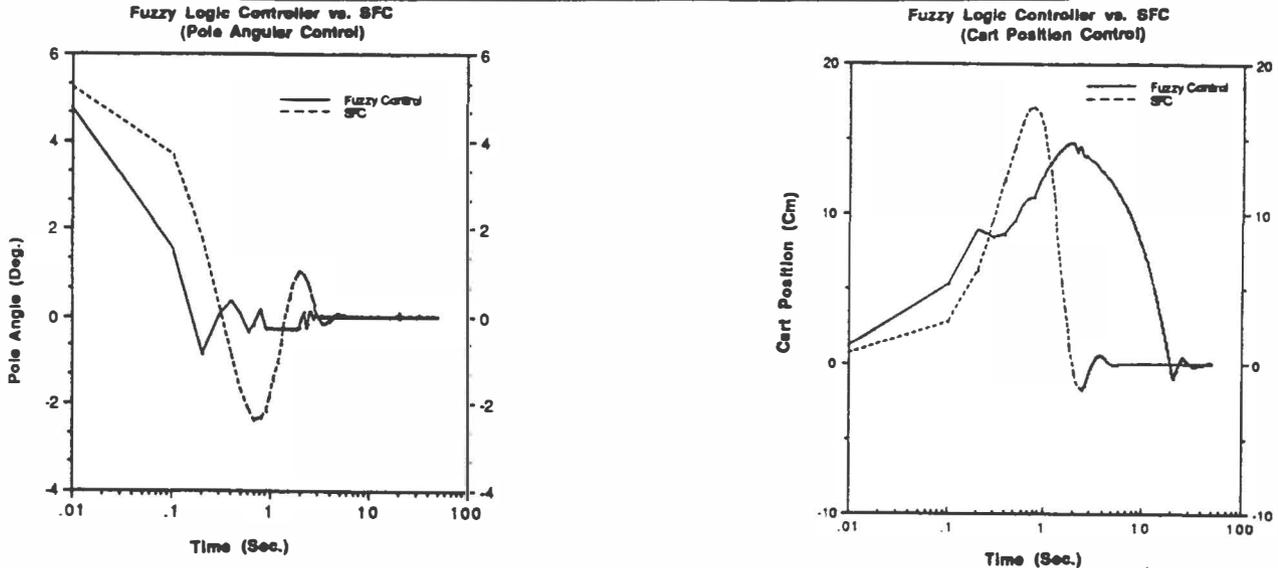

Figure 2: Simulation data - Achievement of goals by the two controllers

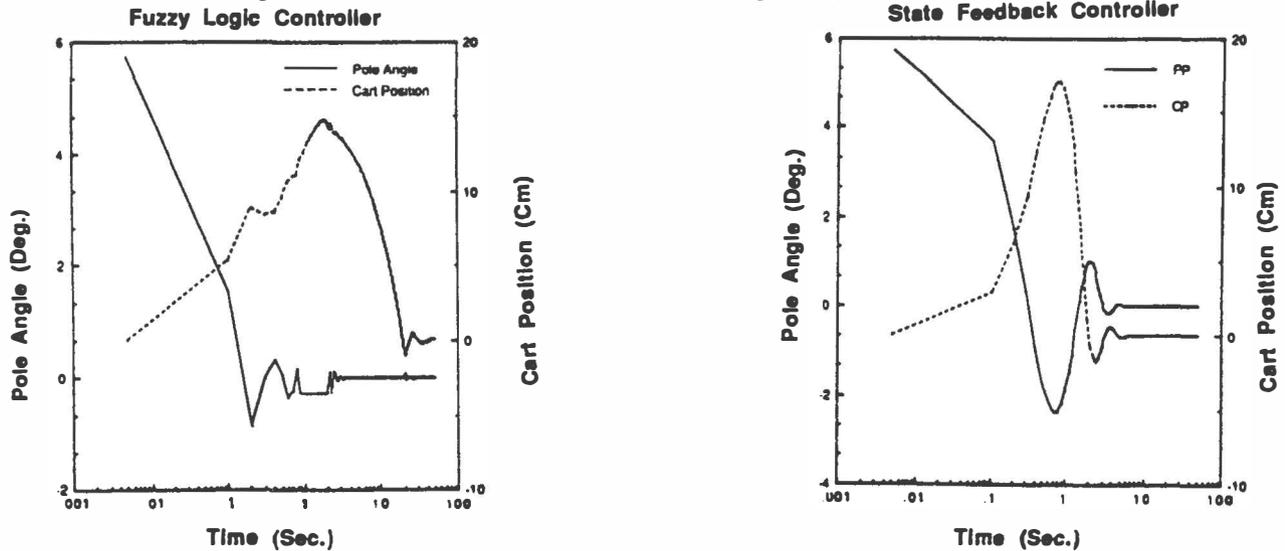

Figure 3: Simulation data - Interaction between the two control goals

In each experiment, we compared the performance of the fuzzy controller with the performance of the state feedback controller. In each case, the fuzzy controller performed better, with less under- and overshoot. However, it took more time for the fuzzy controller to reach stability, especially for controlling the position of the cart. Table 1 summarizes this difference for three of the poles, over a total sampling time of 50 seconds and a simulation time step of 5 mili-seconds.

Figure 2 presents a graphical display of the performance of the fuzzy controller (FC) and the state feedback controller (SFC) in controlling the pole's angular position ($\theta$), and the cart position on the track ($x$). The interactions between these two control goals are shown in Figure 3 for the fuzzy logic controller and the state feedback controller.

**Experiment-Based Comparison:** We implemented both control schemes in a hardware system. The hardware system included an IBM PC-AT, a DC motor, 2 potentiometers for sensing the pole angle and the cart position, and a Data Acquisition and Control Adapter. The cart-pole combination was driven by the DC motor through an aluminum chain which matched the teeth of the driving pulleys (one pulley on the motor shaft). A sampling time of 20 mili-seconds was used.

Figures 4 and 5 illustrate the performance of the hardware system under these two types of control. In our experiment, the fuzzy logic controller succeeded in balancing the pole practically from the beginning of the testing. Some tuning was needed to control the cart position at desired location on the tracks. Not unexpectedly, the SFC controller achieved very good performance also. However, this was accomplished through a long system modeling and identification process (a precise mathematical model had to be acquired before the SFC controller could be implemented).

**Further Experiments:** Several other tests have been performed using the prototype system. Figure 6(a) shows the interaction between pole angular control (goal 1) and cart position control (goal 2) when the pole was tapped twice: once after 15 seconds and once again after 35 seconds. Figure 6 shows the same interaction after the rail tracks were tilted about 7 degrees after 20 seconds and un-tilted after 45 seconds.

We summarize the comparison of these approaches based on the following criteria:

- Design complexity: The design of a fuzzy controller does not require a complete model of the process. SFC is model-based and it requires a precise mathematical model of the process. The complexity of the model was not a problem in our experiment, since analytical models were readily available for the cart-pole balancing problem. However, for a large class of non-linear control problems, this issue is significant.

- Controller modification: Compared to the state feedback controller, a fuzzy logic controller may involve more parameters for fine-tuning. In our experiment, fine-tuning the fuzzy logic controller seemed to be a little more difficult.

- Robustness: The design of the state feedback controller does not include uncertainty in the system, while a fuzzy logic controller, by modeling an operator's knowledge, has a larger tolerance for variations in the process parameters. In our experiments with poles of varying lengths and weights, the fuzzy controller was more robust than the analytical controller. In the case of Pole 7 (i.e., the heaviest and longest pole), the fuzzy controller balanced the pole, albeit with difficulty; the state feedback controller, however, failed to balance the pole.

## Conclusions

We have presented a new method for designing approximate reasoning-based controllers. The hierarchical nature of the method provides a better framework for designers to focus their attention when writing the control rules or fine-tuning them later for smoother performance. We have used POLE and its prototype hardware development to compare the performance of our method with that of an analytical state feedback controller. POLE produced results very close to its counterpart analytical controller and in some cases, POLE's results surpassed them[4]. We believe that these results are good indications of the versatility of our approach in general as a *complement* to the conventional controllers.

Under the assumptions outlined earlier, our method should provide a better facility in designing approximate reasoning-based controllers. Further tests need to be done in applying this approach to other domains. The results reported in this paper have been encouraging enough to start a larger scale project in applying our method to the rendezvous and docking operation of the Space Shuttle with the Space Station or a satellite.

**Acknowledgements:** Many thanks to Peter Friedland, Mark Drummond, and Philip Laird for their valuable comments on earlier draft of this paper.

## Appendix A: Controller's Knowledge Base

| NE | : | Negative       | PL | : | Positive Large  |
|----|---|----------------|----|---|-----------------|
| ZE | : | Zero           | PM | : | Positive Medium |
| PO | : | Positive       | PS | : | Positive Small  |
| VS | : | Very Small     | NS | : | Negative Small  |
| NL | : | Negative Large | NM | : | Negative Medium |

Rules used for angular position control:

Rule-1: IF $\theta$ is PO AND $\dot{\theta}$ is PO THEN F is PL
Rule-2: IF $\theta$ is PO AND $\dot{\theta}$ is ZE THEN F is PM

---

[4] A video, which illustrates the performance of the hardware system, is available from the authors.

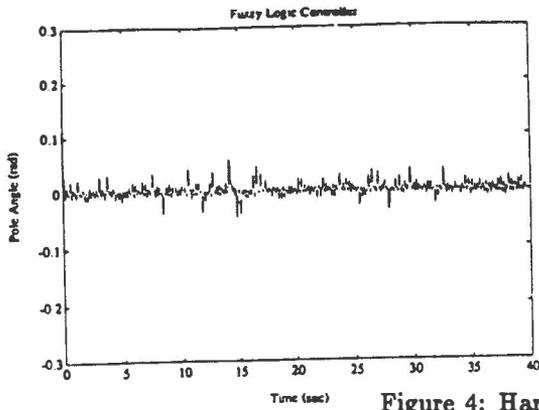
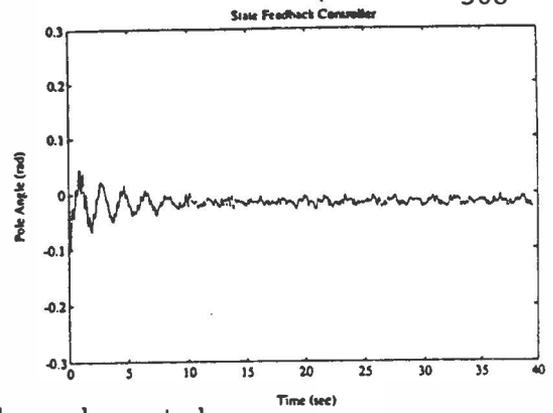

Figure 4: Hardware data - Pole angular control

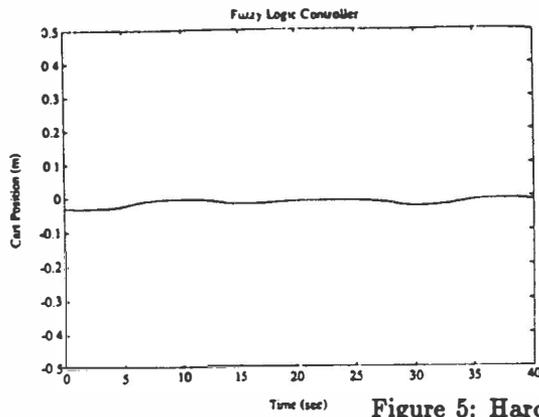
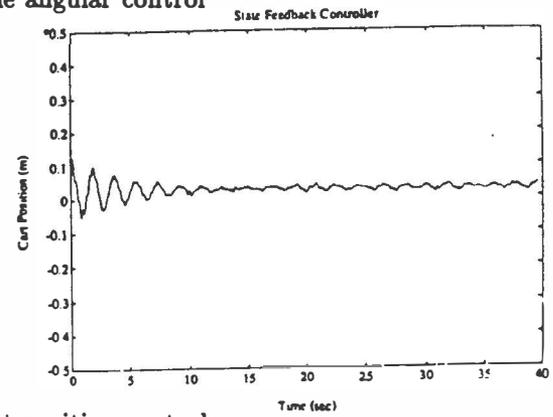

Figure 5: Hardware data - Cart position control

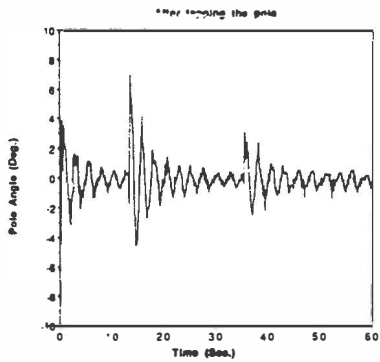
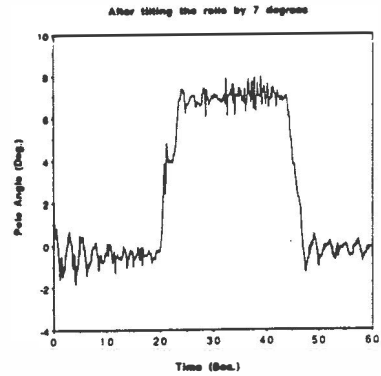
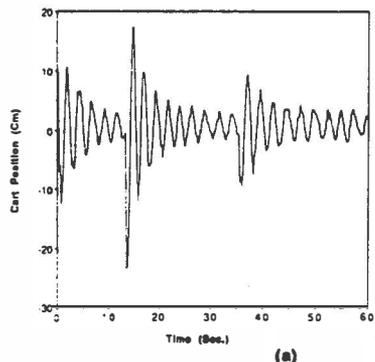
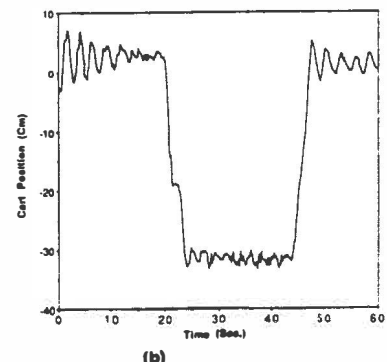

(a)    (b)

Figure 6: Hardware data - Interaction between pole angular control and cart position control After (a) tapping the pole (b) tilting the rail tracks

Rule-3:IF $\theta$ is PO AND $\dot{\theta}$ is NE THEN F is ZE
Rule-4:IF $\theta$ is ZE AND $\dot{\theta}$ is PO THEN F is PS
Rule-5:IF $\theta$ is ZE AND $\dot{\theta}$ is ZE THEN F is ZE
Rule-6:IF $\theta$ is ZE AND $\dot{\theta}$ is NE THEN F is NS
Rule-7:IF $\theta$ is NE AND $\dot{\theta}$ is PO THEN F is ZE
Rule-8:IF $\theta$ is NE AND $\dot{\theta}$ is ZE THEN F is NM
Rule-9:IF $\theta$ is NE AND $\dot{\theta}$ is NL THEN F is NL

Rules used for horizontal position control of the cart:

Rule-10:IF $\theta$ is VS AND $\dot{\theta}$ is VS AND $x$ is PO AND $\dot{x}$ is PO THEN F is PM
Rule-11:IF $\theta$ is VS AND $\dot{\theta}$ is VS AND $x$ is PO AND $\dot{x}$ is ZE THEN F is PS
Rule-12:IF $\theta$ is VS AND $\dot{\theta}$ is VS AND $x$ is NE AND $\dot{x}$ is NE THEN F is NM
Rule-13:IF $\theta$ is VS AND $\dot{\theta}$ is VS AND $x$ is NE AND $\dot{x}$ is ZE THEN F is NS

### Appendix B: Cart-pole balancing governing equations

The governing equations of the cart-pole balancing problem are given by the following nonlinear differential equations [Barto 83]:

$$\ddot{\theta} = \frac{g \sin\theta + \cos\theta\left[\frac{-f - ml\dot{\theta}^2 \sin\theta + \mu_c sgn(\dot{x})}{m_c + m}\right] - \frac{\mu_p \dot{\theta}}{ml}}{l\left[\frac{4}{3} - \frac{m\cos^2\theta}{m_c + m}\right]}$$

$$\ddot{x} = \frac{f + ml[\dot{\theta}^2 \sin\theta - \ddot{\theta}\cos\theta] - \mu_c sgn(\dot{x})}{m_c + m}$$

where $\theta$ is the angle of the pole with respect to the vertical line, $x$ is the horizontal position of the cart, $f$ is the driving force applied to the cart, $g$ is the acceleration due to gravity, $m_c$ is the mass of the cart, $m$ is the mass of the pole, $l$ is the half-pole length, $\mu_c$ is the coefficient of friction of cart on track, and $\mu_p$ is the coefficient of friction of pole on cart.